\documentclass[10pt]{article}
\usepackage{blindtext}
\usepackage{natbib}
\usepackage{setspace}
\usepackage{epigraph}
\usepackage{caption}
\usepackage{ragged2e}
\usepackage{csquotes} 
\usepackage{pdflscape}
\usepackage{amsmath}
\usepackage{amssymb} 
\usepackage{setspace}
\usepackage{footmisc}
\usepackage{mathtools}

\usepackage{enumitem}
\usepackage{amsmath}
\usepackage{epigraph}
\usepackage{lipsum}
\usepackage{authblk}
\usepackage[labelfont=bf, font=footnotesize]{caption}
\usepackage{bibunits}
 
\usepackage[dvipsnames]{xcolor}
\usepackage{lineno}
\usepackage{titlesec}
\makeatletter
\def\thanks#1{\protected@xdef\@thanks{\@thanks
        \protect\footnotetext{#1}}}
\makeatother
\titleformat*{\section}{\Large\bfseries}

\usepackage{graphicx}
\usepackage[colorlinks=true,linkcolor=Blue, urlcolor  = Blue, citecolor=Blue]{hyperref}%
\usepackage{geometry}
\title{\vspace{-.25in}
\LARGE \textbf{On the Unknowable Limits to Prediction}\vspace{0.35in}
}

\author[1,2,3,*]{\large{Jiani Yan\thanks{\noindent 
 \textbf{For Correspondence:} Jiani Yan and Charles Rahal, Leverhulme Centre for Demographic Science, Demographic Science Unit, University of Oxford, OX1 1JD, United Kingdom. Tel: 01865 286170. Email: \href{mailto:jiani.yan@sociology.ox.ac.uk}{jiani.yan@sociology.ox.ac.uk} and \href{mailto:charles.rahal@demography.ox.ac.uk}{charles.rahal@demography.ox.ac.uk}. \textbf{Acknowledgements}: Funding is gratefully acknowledged from the Leverhulme Trust (Grant RC-2018-003) for the Leverhulme Centre for Demographic Science, Nuffield College, and the Economic and Social Research Council. Comments gratefully received from Kyla Chasalow, Mark Verhagen, Ridhi Kashyap, Ben Domingue, Michael Biggs, Sander Wagner, Haohao Lei and Edith Darin. We are especially grateful to comments received from Nicholas Irons. A GitHub repository which replicates the figures in this manuscript is available at \href{https://github.com/crahal/unpredictability}{github.com/crahal/unpredictability}. This is indexed into Zenodo as DOI: 10.5281/zenodo.14674892. \href{https://zenodo.org/records/14674893}{(Rahal and Yan, 2025)}}}}
\author[1,4,*]{\large{Charles Rahal}}
\affil[1]{\small{Leverhulme Centre for Demographic Science, University of Oxford}}
\affil[2]{\small{Department of Sociology, University of Oxford}}
\affil[3]{\small{Wolfson College, University of Oxford}}
\affil[4]{\small{Nuffield College, University of Oxford}}
\affil[*]{\small{Both authors contributed jointly to this work.}}

\date{\vspace{0in}\today\vspace{0.2in}}

\begin{document}
\maketitle
\begin{bibunit}[chicago]

\newpage

\setlength{\epigraphwidth}{0.625\textwidth}
 \epigraph{%
    \begin{minipage}{\epigraphwidth}
     \justifying
     ``\textit{We may regard the present state of the universe as the effect of its past and the cause of its future. An intellect which at a certain moment would know all forces that set nature in motion, and all positions of all items of which nature is composed, if this intellect were also vast enough to submit these data to analysis, \dots nothing would be uncertain.}''%
     \end{minipage}
 }{-- \cite{laplace1814}, \textit{Essai philosophique sur les probabilit\'{e}s}}

\vspace{0.1in}
 The classic dichotomisation of prediction error into statically defined `reducible' and `irreducible' terms is unable to capture the intricacies of progressive research design where specific types of \emph{truly} reducible error are being rapidly eliminated at differential speeds. While this is generalisably true -- including in the domain of Large Language Models (LLMs) -- it is none more apparent than in social systems where questions regarding the `predictability' of outcomes are gradually reaching the fore \citep{rahal2024rise}. Canonical work which convenes common predictive tasks involving conventional social surveys \citep{salganik2020measuring} shows low power, but rapidly emerging computational approaches which utilize non-standard administrative data and information contained within things such as children's autobiographical essays -- often combined with generative Artificial Intelligence and distributed computing \citep{wolfram2022short,savcisens2024using} -- begin to allow high accuracy. We propose an enhanced framework which builds upon existing work \citep{hullermeier2021aleatoric} in a way that we believe helpfully decomposes various types of truly reducible `epistemic' error which are in theory eliminable (resulting from a lack of knowledge), and isolates residual, `aleatoric' error (inherent randomness that can \emph{never} be modeled) that research endeavor will not be able to capture in the limit of human progress. We further emphasize the impossibility of claiming whether things are `predictable', especially with regards to open, dynamically evolving systems. 
 
Decomposing prediction error into reducible and irreducible terms is not new.\footnote{See, for example, Sections 2.9 and 7.3 of \cite{hastie2009elements}; a landmark instruction manual in the field. 
This foundational text uses the term `reducible' to taxonomically classify errors into a type that can be eliminated, and the term `irreducible' to denote errors which cannot be eliminated \emph{given current models and information on target variables and feature sets.}}
 However, statements regarding `predictability' and `irreducibility' require the reinforcement of an important quantifier; they are entirely conditional on information sets. For example, the insightful mixed-methods work of \cite{lundberg2024origins} begins its abstract with `Why are some life outcomes difficult to predict?'. In that case, they are difficult to predict based on information that is currently measurable, and the `task' at hand. An outcome which is difficult to predict with one feature set may not be `unpredictable' in an aleatoric sense: life trajectories may be trivial to predict in the future. Reflexive terminology is essential, lest conclusions be drawn that there is no space for future positive policy interventions based on ever increasingly accurate predictions that prevent negative outcomes. 
It is impossible to know whether we have eliminated all reducible epistemic error to approach the practical `ceiling' of accuracy, make statements about how predictable outcomes will become, or understand where we are in the process. 
Predictability of an outcome can only be asserted when all epistemic errors are eliminated: when `unmeasured' information is measured and constructed perfectly, features are constructed perfectly, and when functional forms are able to be exactly recovered.\par

Let $y_{\text{true}}$ denote the actual, observable outcome that is measured after constructing the phenomenon of interest; $\varepsilon$ is aleatoric error for (that specific) $y_{\text{true}}$. Next, denote $\mathbf{x}_{\text{true}}$ as the perfectly chosen, constructed and measured feature set from a possibly uncountable, infinite collection of
features in the universe, and $f^*()$ as the conceptually true underlying function that maps $\mathbf{x}_{\text{true}}$ onto $y_{\text{true}}$. As opposed to the scenario of perfect prediction described above, $y_{\text{observed}},\mathbf{x}_{\text{observed}} \text{ and } f(\mathbf{x}| y)$ are defined as the observed target, features, and the model trained on them. Equation \ref{eq:example} expresses our suggestion for conceptually decomposing predictive error under the assumption of no distributional shift: 

\begin{align} \label{eq:example}
    y_{\text{true}} &= \underbrace{f^*(\mathbf{x}_{\text{true}})}_{\substack{\text{Predictive} \\ \text{Ceiling}}} + \underbrace{\varepsilon}_{\substack{\text{Aleatoric} \\ \text{Error}}} \\
        &= \underbrace{[f^*(\mathbf{x}_{\text{true}}) - f(\mathbf{x}_{\text{true}} \mid y_{\text{true}})]}_{\substack{\text{Model approximation gain} \\ \text{(Epistemic)}}} \nonumber \\
        &\quad+ \underbrace{[ f(\mathbf{x}_{\text{true}} \mid y_{\text{true}}) - f(\mathbf{x}_{\text{true}} \mid y_{\text{observed}})]}_{\substack{\text{Measurement gain from } y \\ \text{(Epistemic)}}} \nonumber \\
        &\quad+ \underbrace{[ f(\mathbf{x}_{\text{true}} \mid y_{\text{observed}}) - f(\mathbf{x}_{\text{observed}} \mid y_{\text{observed}})]}_{\substack{\text{Measurement gain from } \mathbf{x} \\ \text{(Epistemic)}}} \nonumber \\
        &\quad+ \underbrace{y_{\text{predicted}}}_{\substack{\text{Current Prediction}}} + \underbrace{\varepsilon}_{\substack{\text{Irreducible} \\ \text{(Aleatoric)}}}. \nonumber
\end{align}

\begin{figure}[!t]
\begin{center}
\includegraphics[width=1\textwidth]{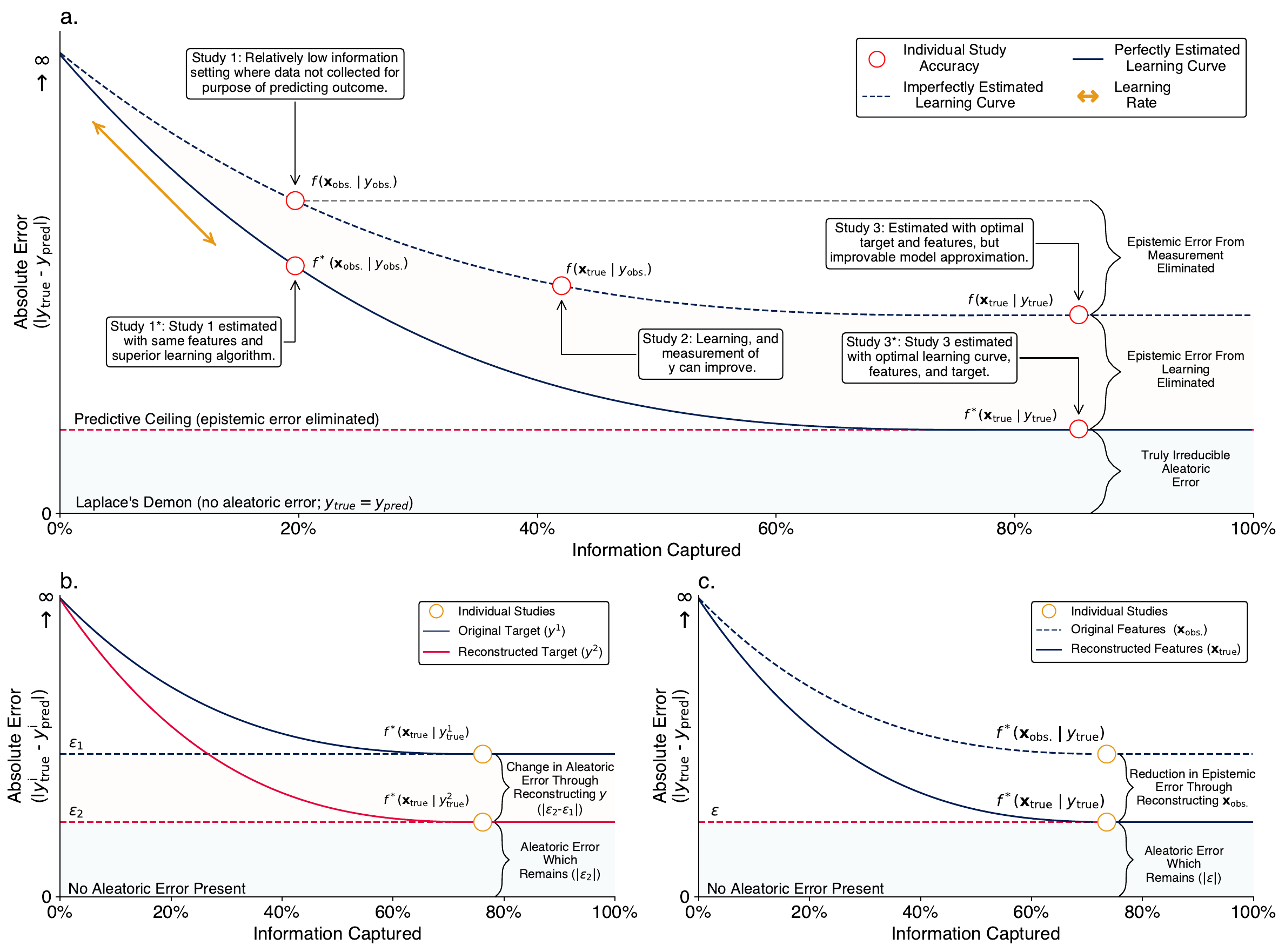}
\caption{\textbf{Decomposed Learning Curves.} The functional form of learning curves can vary depending on the targets, algorithms, and features being used, and different amounts of truly irreducible error may exist for different target variables (see Supplementary Figure \ref{figure_si1}). As relevant information accumulates, the learning curve -- defined as the functional representation of a model's predictive performance against the amount of information which it receives -- is expected to be monotonic, but may not be entirely continuous.  Panel `a.' represents a baseline scenario which positions specific studies against how well information on target and features has been measured (and how well such information has been mapped into learning algorithms), while Panels `b.' and `c.' represent changes (and gains and changes in predictive performance) with regards to construct validity of $y$ and $\mathbf{x}$ respectively. Note: `obs.' denotes `observed' variables as per Equation \ref{eq:example}.}
\label{figure_1}
\end{center}
\end{figure}

\noindent As $\mathbf{x}_{\text{true}}$  and $y_{\text{true}}$ are enhanced information sets over and above $\mathbf{x}_{\text{observed}}$ and $y_{\text{observed}}$, they could tautologically reduce what may currently have been termed `irreducible'.\footnote{For class labeling tasks (such as binary classification), our conceptual framework remains viable at the aggregate level with appropriately defined model evaluation metrics: we leave this -- as well as extensions to things such as probability distributions, entropy, and stochastic processes adapted to filtration -- for further work.} Predictive accuracy could also improve through better construct validity \citep{o1998empirical}: the extent to which the model's predictions or features accurately capture the theoretical construct or concept the model is intended to measure or represent.\footnote{Note: better construction of the target and features may not necessarily involve the collection of additional information, but simply a re-framing of a variable or new utilization of existing information, as depicted in Figures \ref{figure_1}b.-c.} Incorporation of previously unmeasured information (from $\mathbf{x}_{\text{observed}}$ towards $\mathbf{x}_{\text{true}}$) or better measurement of existing features (from $\mathbf{x}_{\text{observed}}$ and $y_{\text{observed}}$ towards $\mathbf{x}_{\text{true}}$ and $y_{\text{true}}$) will also reduce epistemic error, as will reductions in model approximation error (from $f \text{ towards }f^*$).\footnote{Kindly see our Supplementary Information for a fully delineated decomposition of Equation \ref{eq:example}.} To illustrate how these different forms of error reduction work in practice, we depict our generalized framework in Figure \ref{figure_1}. This demonstrates how learning curves evolve as different sources of epistemic errors are reduced (Figure \ref{figure_1}a.) and targets and features are constructed differently (Figure \ref{figure_1}b.-c.).\footnote{Supplementary Information \ref{figure_si1}a. shows the case where aleatoric error may be low and the speed of improvement relative to information gain is high with large gains from measurement error possible. Supplementary Information Figure \ref{figure_si1}b. shows the case where aleatoric error may be high, but learning is relatively slower and only minimal gains from reducing learning error are possible.} For ease of exposition, Equation \ref{eq:example} represents the prediction of one out-of-sample value, and an aggregate level decomposition can be found in the Supplementary Information.

The existence of aleatoric errors means we cannot appreciate a practical limit to prediction, because that itself cannot be measured. We have no reasonable expectation of what data will be collectible, or how powerful deep learning algorithms may become. LLMs -- as a specific class of these algorithms -- 
exemplify how scaling laws suggest accuracy improves systematically with model size and data, 
probing the boundaries of epistemic and aleatoric errors. 
It is only recently that the concept of measuring epigenetic data -- let alone interpreting it at scale -- has become viable due to advances in molecular biology. Aleatoric error may be large and nontrivial, with suggestions that predictive performance may be bounded well below deterministic precision. Some data may be potentially impossible to collect, and some outcomes may be genuinely stochastic. The `time to event' will also affect how large aleatoric error is; it is easier to predict how somebody will vote at the next election than how the same person will vote in three election's time. Similarly, re-framing the outcome from `when will somebody have a child' to `whether somebody will have a child' will also affect the size of aleatoric error. We may be able to better approximate `perfect prediction' in some easily constructed and perfectly measurable outcomes (for instance, text prediction and machine translation) than others (for instance, behavioral traits). Research is required to perturb information sets across individual prediction exercises (for instance, through the comparison of better measurement and construction with differential inclusion/omission), illuminating how quickly each type of epistemic error is reduced. Work on systematically indexing predictive accuracy is also of utility, as are designs which seek to maximize external validity. Emerging qualitative approaches which explore predictive errors as a function of pivotal moment's in people's lives \citep{lundberg2024origins} are also immensely promising, and represent a more nuanced form of information. Collecting more detailed information also necessitates careful discussions around privacy, compliance, and ethical issues, particularly as LLMs often rely on vast datasets that may inadvertently include proprietary or copyrighted material, raising concerns about data provenance, consent, and the potential for intellectual property violations. However, until we have reason to believe that we have eliminated error through better measurement and construction of features and outputs -- in addition to the elimination of learning error --  we should only tentatively discuss how predictable things are with current data and technology.


\label{sec:Bibliography}
\end{bibunit}

\newpage
\setcounter{figure}{0}
\setcounter{section}{0}
\setcounter{equation}{0}

\begin{bibunit}[chicago]

\linenumbers
\section{Supplementary Information}\label{si_decomp}

\subsection{Delineating Epistemic and Aleatoric Errors}

In predictive modeling, researchers often aim to predict a phenomenon or concept of interest; they aim to construct a concrete and measurable representation of this as $y_{\text{true}}$. However, due to measurement errors that arise from factors such as subjective assessment, accounting inaccuracies, or sampling bias, researchers are often forced to conceptualize this phenomenon which they want to measure as something similar to $y_{\text{observed}}$. Such measurement errors exist in both feature and target variables; it is essential to understand how these errors propagate through the model and affect predictive accuracy. This supplementary section presents a mathematical framework to link $y_{\text{true}}$ with $y_{\text{observed}}$, highlighting epistemic (reducible) and aleatoric (irreducible) errors in prediction, and then equates it to the familiar bias-variance trade-off.

\subsubsection{Problem Setup}\label{sec:problem}

Consider the following definition of the `true' target and features, the observable target and features, and the model which binds them as follows.\\

\noindent\noindent\textit{1. True Target and Features}\\

\noindent Allow us to equate the relationship between the `true' target variable and feature set as: 
\begin{equation}
    y_{\text{true}} \coloneqq f^*(\mathbf{x}_{\text{true}}) + \varepsilon,
    \label{eq:true_outcome}
\end{equation}
\noindent where:
\begin{itemize}
    
    \item $y_{\text{true}}$ is the true outcome, measured \emph{after} constructing the phenomenon of interest;
    
    \item $\mathbf{x}_{\text{true}}$ represents the best possible feature set both in terms of quantity and in quality. This is a perfectly chosen, constructed and measured feature set;
    
    \item $f^*()$ is the conceptually true underlying function which maps $\mathbf{x}_{\text{true}}$ onto $y_{\text{true}}$, which can be specifically written as $f^*: \mathbb{R}^n \to \mathbb{R}, \ f^*(\mathbf{x}_{\text{true}}) = y_{\text{true}}-\varepsilon$;
    
    \item $\varepsilon$ is aleatoric error (inherent randomness) specific to $y_{\text{true}}$, with $\mathbb{E}[\varepsilon] = \varepsilon$ and variance $\operatorname{Var}(\varepsilon) = \sigma_{\varepsilon}^2$. It is fixed upon the definition of $y_{\text{true}}$, assuming that statistical properties of the target do not change over time or across different contexts (i.e., no distributional shift). However, in the real-world, such an assumption may not always hold, causing the level of $\varepsilon$ to fluctuate over time.\\
\end{itemize}

\noindent\noindent\textit{2. Observed Target and Features}\\

\noindent Next, allow us to introduce measurement error in $y_{\text{true}}$ and $\mathbf{x}_{\text{true}}$ by relating them to what we can observe:
\begin{align}
    y_{\text{observed}} &\coloneqq y_{\text{true}} + \delta_y  = f^*(\mathbf{x}_{\text{true}}) + \varepsilon + \delta_y , \label{eq:observed_outcome} \\
    \mathbf{x}_{\text{observed}} &\coloneqq \mathbf{x}_{\text{true}} + \delta_\textbf{x}, \label{eq:observed_predictors}
\end{align}

\noindent where:
\begin{itemize}
\item $\delta_y$ is measurement error in $y$, with $\mathbb{E}[\delta_y] = \mu_{\delta_y}$ and $\operatorname{Var}(\delta_y) = \sigma_{\delta_y}^2$;

\item $\delta_\mathbf{x}$ is measurement error in $\mathbf{x}$, with $\mathbb{E}[\delta_\mathbf{x}] = \mu_{\delta_\mathbf{x}}$ and variance $\operatorname{Var}(\delta_\mathbf{x}) = \sigma_{\delta_\mathbf{x}}^2$. $\mathbf{x}_{\text{observed}}$ is a vector with $1 \times K$ dimensions, with each $x_k \in \mathbf{X}$ for $k=\{1, 2, ..., K\}$ for each one of the predictors. Note: $\mathbf{x}_{\text{observed}}$ can have different dimensions to $\mathbf{x}_{\text{true}}$. 

\end{itemize}
$\delta_\textbf{x}$ can be reduced through improved construction of each $x_k \in \mathbf{X}$; increased sample representativeness, and conceptualizing the feature set so that it contains the most relevant set of predictors. In general, we have assumed that $y_{\text{observed}}$ captures all aleatoric errors in $y_{\text{true}}$; changing $\delta_y$ does not impact the level of aleatoric error $\varepsilon$. A discussion of the violation of this condition can be found in Supplementary Information Section \ref{sec: partial_aleatoric} \\

\noindent\noindent\textit{3. Predictive Model}\\

\noindent The predictive model is estimated using observable training data $\textbf{x}_{\text{observed}}$ and $y_{\text{observed}}$:

\begin{equation}
    y_{\text{pred}} \coloneqq f(\textbf{x}_{\text{observed}} \mid y_{\text{observed}}), \label{eq:predictive_model}
\end{equation}

\noindent where ${f}$ is the estimated model trained on $\{\textbf{x}_{\text{observed}}, y_{\text{observed}}\}$. The notation ${f}(\textbf{x}_{\text{observed}} \mid y_{\text{observed}})$ emphasizes that ${f}$ depends on $y_{\text{observed}}$ in training processes. The difference between the conceptually true underlying function $f^*()$ and the currently selected predictive function $f()$ on $\textbf{x}_{\text{true}}$ and $y_{\text{true}}$ is defined as follows:

\begin{equation}
    f(\textbf{x}_{\text{true}} \mid y_{\text{true}})  \coloneqq f^*(\textbf{x}_{\text{true}}) + \delta_f, 
    \label{eq:model_approximation_error}
\end{equation}

\noindent where $\delta_f$ is defined based on the optimal $\textbf{x}_{\text{true}}$ and $y_{\text{true}}$, and $\mathbb{E}[\delta_f] = \mu_{\delta_f}, \operatorname{Var}[\delta_f] = \sigma^2_{\delta_f}$. The difference between $f(\textbf{x}_{\text{true}} \mid y_{\text{true}})$ and $y_{\text{pred}}$ attributes to improvement in $\delta_\textbf{x}$ and $\delta_y$; further details can be found in Supplementary Information Equation \ref{eq:y_pred_decomposition_refined}.

\subsubsection{Decomposing $y_{\text{true}}$}
Following these aforementioned definitions, we can decompose the output of a machine which predicts $y_{\text{true}}$ as perfectly as possible, and in the process link it to existent predictions generated by $f()$ to create $y_{\text{pred}}$ as follows:

\begin{align}
        y_{\text{true}} &=   f^*(\textbf{x}_{\text{true}}) + \varepsilon \nonumber \\
        &=  [f^*(\textbf{x}_{\text{true}}) - f(\textbf{x}_{\text{true}} \mid y_{\text{true}})]  \nonumber \\
        &\quad+  [ f(\textbf{x}_{\text{true}} \mid y_{\text{true}}) - f(\textbf{x}_{\text{true}}\mid y_{\text{observed}})] \nonumber \\
        & \quad+  [ f(\textbf{x}_{\text{true}}\mid y_{\text{observed}}) - f(\textbf{x}_{\text{observed}}\mid y_{\text{observed}})] \nonumber \\
        & \quad+  f(\textbf{x}_{\text{observed}}\mid y_{\text{observed}}) + \varepsilon \nonumber \\
        &= \underbrace{ -\delta_f}_{\shortstack{\scriptsize\text{Model approximation gain} \\ \scriptsize\text{(Epistemic)}}} \nonumber \\
        &\quad+ \underbrace{[ f(\textbf{x}_{\text{true}} \mid y_{\text{true}}) - f(\textbf{x}_{\text{true}} \mid y_{\text{true}}+\delta_y)]}_{\scriptsize\shortstack{\text{Measurement gain from} y \\ \text{(Epistemic)}}} \nonumber \\
        & \quad+  \underbrace{[f(\textbf{x}_{\text{true}} \mid y_{\text{observed}}) - f(\textbf{x}_{\text{true}} + \delta_\textbf{x} \mid y_{\text{observed}})]}_{\scriptsize\shortstack{\text{Measurement gain from} $\mathbf{x}$ \\ \text{(Epistemic)}}}  \nonumber \\
        & \quad+  \underbrace{y_{\text{pred}}}_{\scriptsize\shortstack{\text{Current Prediction}}} + \underbrace{\varepsilon}_{\scriptsize\shortstack{\text{Irreducible}\\ \text{(Aleatoric)}}}. \nonumber \\
        \label{eq:y_pred_decomposition_refined}
\end{align}

\noindent Therefore, we can achieve improved predictive performance in comparison to existing predictions based on $\textbf{x}_{\text{observed}}$, $y_{\text{observed}}$, and $f()$. This is made possible through opportunities to eliminate measurement and construction error in $\textbf{x}_{\text{observed}}$ (i.e., reducing $\delta_\textbf{x}$), measurement in $y_{\text{observed}}$ (i.e., reducing $\delta_y$), and better model approximation $f()$ (i.e., reducing $\delta_f$). When $\delta_\textbf{x} = \delta_y = \delta_f = 0$, we reach a scenario where the difference between $y_{\text{true}}$ and $y_{\text{predicted}}$ is $\varepsilon$: pure aleatoric error. This essentially represents the `upper bound' of predictive accuracy, occasionally referred to as a `predictive ceiling' \citep{garip2020failure}.

\subsubsection{Error Decomposition}

For any individual observation, the total prediction error (denoted below as $Error$) is defined as the difference between the predicted and true outcome:

\begin{align}
    Error &\coloneqq y_{\text{pred}} - y_{\text{true}}  \nonumber \\
    &= {f}(\textbf{x}_{\text{observed}} \mid y_{\text{observed}}) - y_{\text{true}} \nonumber \\
    &= \left[ {f}(\textbf{x}_{\text{observed}} \mid y_{\text{observed}}) - {f}(\textbf{x}_{\text{true}} \mid y_{\text{observed}}) \right] \nonumber \\
    &\quad + \left[ {f}(\textbf{x}_{\text{true}} \mid y_{\text{observed}}) - {f}(\textbf{x}_{\text{true}} \mid y_{\text{true}}) \right] \nonumber \\
    &\quad + \left[ {f}(\textbf{x}_{\text{true}} \mid y_{\text{true}}) - f^*(\textbf{x}_{\text{true}}) \right] + \left[ f^*(\textbf{x}_{\text{true}}) - y_{\text{true}} \right]. \nonumber \\
    \label{eq:total_error}
\end{align}

\noindent Substituting for Supplementary Information Equations \ref{eq:true_outcome}-\ref{eq:observed_predictors}, we obtain: 

\begin{align}
    Error &= \underbrace{\left[ {f}(\textbf{x}_{\text{observed}} \mid y_{\text{observed}}) - {f}(\textbf{x}_{\text{observed}}-\delta_\textbf{x} \mid y_{\text{observed}}) \right]}_{\shortstack{\scriptsize\text{Error due to measurement error in $\mathbf{x}$} \\ \\\scriptsize\text{(Epistemic)}}} \nonumber \\    
    &\quad + \underbrace{\left[ {f}(\textbf{x}_{\text{true}} \mid y_{\text{observed}}) - {f}(\textbf{x}_{\text{true}} \mid y_{\text{observed}}-\delta_y ) \right]}_{\shortstack{\scriptsize\text{Error due to measurement error in $y$ affecting model estimation}\\ \scriptsize\text{(Epistemic)}}} \nonumber \\
    &\quad + \quad\quad\quad \underbrace{\delta_f}_{\shortstack{\scriptsize\text{Model approximation error}\\ \scriptsize\text{(Epistemic)}}}\quad+ \underbrace{\varepsilon}_{\shortstack{\scriptsize\text{Irreducible}\\ \scriptsize\text{(Aleatoric)}}}. \nonumber \\
    \label{eq:error_decomposition}
\end{align}

\noindent It then logically follows that predicting the target variable as accurately as can be done in practice (see Supplementary Information Section \ref{sec:problem}) results in the case where all epistemic errors are eliminated, and only aleatoric error remains: $\text{Error} = y_{\text{pred}}-y_{\text{true}}=\varepsilon$.

\subsubsection{Relationship to Existing Frameworks}


Our framework mainly decomposes errors in a conceptually simple way (i.e., up until now, we have been dealing with single scalar target outcome observations). We next develop the aggregate level decomposition of the expected sum of squared errors and relate our framework to the bias, variance, and error frameworks of canonical textbooks (e.g., \citealp{james2013introduction, hastie2009elements}) as well as papers (\citealp{pedro2000unified, james2003variance}) in the field of statistical learning. Suppose we have a sample with $N$ observations, with each $y_{\text{true},i} \in \textbf{y}_{\text{true}}$ and $y_{\text{pred},i} \in \textbf{y}_{\text{predicted}}$ for $i=\{1, 2, ..., N\}$. Supplementary Information Equation \ref{eq:error_decomposition} holds as follows: 

\begin{align}
    Error &= y_{\text{pred}}-y_{\text{true}} \nonumber \\
    &= \underbrace{\left[ {f}(\textbf{x}_{\text{observed}} \mid y_{\text{observed}}) - {f}(\textbf{x}_{\text{observed}}-\delta_\textbf{x} \mid y_{\text{observed}}) \right]}_{\shortstack{\scriptsize\text{Error due to measurement error in $\mathbf{x}$} \\ \\\scriptsize\text{(Epistemic)}}} \nonumber \\    
    &\quad + \underbrace{\left[ {f}(\textbf{x}_{\text{true}} \mid y_{\text{observed}}) - {f}(\textbf{x}_{\text{true}} \mid  y_{\text{observed}}-\delta_y ) \right]}_{\shortstack{\scriptsize\text{Error due to measurement error in $y$ affecting model estimation}\\ \scriptsize\text{(Epistemic)}}} \nonumber \\
    &\quad + \quad\quad\quad \underbrace{\delta_f}_{\shortstack{\scriptsize\text{Model approximation error}\\ \scriptsize\text{(Epistemic)}}}\quad+ \underbrace{\varepsilon}_{\shortstack{\scriptsize\text{Irreducible}\\ \scriptsize\text{(Aleatoric)}}} \nonumber \\
    &= \delta^{\text{error}}_\mathbf{x} + \delta^{\text{error}}_y + \delta_f+ \varepsilon,
\end{align}
\noindent where:
\begin{itemize}

    \item $\delta^{\text{error}}_\textbf{x} \coloneqq  {f}(\textbf{x}_{\text{observed}} \mid y_{\text{observed}}) - {f}(\textbf{x}_{\text{observed}}-\delta_\textbf{x}\mid y_{\text{observed}})$; the difference in predicted values owing to $\delta_\textbf{x}$. $\mathbb{E}[\delta^{\text{error}}_\textbf{x}] = \mu^{\text{error}}_\textbf{x}$ and variance $\operatorname{Var}(\delta^{\text{error}}_\textbf{x}) = \sigma^2_{\delta^{\text{error}}_\textbf{x}}$. As $\delta^{\text{error}}_\textbf{x}$ is expressed as the difference between the fitted values, the calculation of $\operatorname{Var}(\delta^{\text{error}}_\textbf{x})$ depends on the specific form of $f$, the distribution of $\textbf{x}_{\text{observed}} \text { and } \textbf{x}_{\text{observed}}-\delta_\textbf{x}$, and their dependencies;
    
    \item $\delta^{\text{error}}_y \coloneqq  {f}(\textbf{x}_{\text{true}} \mid y_{\text{observed}}) - {f}(\textbf{x}_{\text{true}} \mid  y_{\text{observed}}-\delta_y)$; the difference in predicted values owing to $\delta_y$. $\mathbb{E}[\delta^{\text{error}}_y] = \mu^{\text{error}}_y$ and variance $\operatorname{Var}(\delta^{\text{error}}_y) = \sigma^2_{\delta^{\text{error}}_y}$. Similarly, the calculation of $\operatorname{Var}(\delta^{\text{error}}_y)$ depends on the specific form of $f$, the distribution of $y_{\text{observed}} \text { and } y_{\text{observed}}-\delta_y$, and their dependencies;

    \item $\delta_f$ is defined as before in Equation \ref{eq:model_approximation_error}.
\end{itemize}

\noindent Therefore, 
\begin{align}
    \mathbb{E}[Error] &= \mathbb{E}[\delta^{\text{error}}_\textbf{x} + \delta^{\text{error}}_y + \delta_f+ \varepsilon] \nonumber \\
    &= \mu^{\text{error}}_\textbf{x} + \mu^{\text{error}}_y + \mu_{f}+\varepsilon.
    \label{eq:bias_expectation}
\end{align}

\noindent Since $\varepsilon$ is uncorrelated with other sources of error:

\begin{align}
    \operatorname{Var}(Error) &= \operatorname{Var}[(\delta^{\text{error}}_\textbf{x} + \delta^{\text{error}}_y + \delta_f+ \varepsilon)] \nonumber \\
    &=  \operatorname{Var}[(\delta^{\text{error}}_\textbf{x} + \delta^{\text{error}}_y + \delta_f)]+\operatorname{Var}(\varepsilon).
    \label{eq:bias_var}
\end{align}

\noindent With Supplementary Information Equations \ref{eq:bias_expectation}-\ref{eq:bias_var}, we can write our squared error loss as: 
\begin{align}
    \mathbb{E}[(y_\text{true} - y_\text{pred})^2] &= \mathbb{E}[(y_{\text{true}}-y_{\text{pred}})]^2 + \operatorname{Var}[(y_{\text{true}}-y_{\text{pred}})] \nonumber \\
    & = \mathbb{E}[Error]^2 + \operatorname{Var}(Error) \nonumber \\ 
    & = (\underbrace{\mu^{\text{error}}_\textbf{x} + \mu^{\text{error}}_y + \mu_{\delta_f}+\varepsilon}_{\text{Bias}})^2 + \underbrace{\operatorname{Var}[(\delta^{\text{error}}_\textbf{x} + \delta^{\text{error}}_y + \delta_f)+]}_{\text{Variance}} + \underbrace{\operatorname{Var}(\varepsilon)}_{\text{Aleatoric Error}}, \label{eq:bias_var_decomposition}
\end{align}
\noindent which results in the familiar $\text{Irreducible Error} + \text{Bias}^2 + \text{Variance}$ framework akin to Equation 7.9 of \cite{hastie2009elements}. For simplicity, we don't expand $\operatorname{Var}[(\delta^{\text{error}}_\textbf{x} + \delta^{\text{error}}_y + \delta_f)]$, which would include the covariance between each of the $\delta$ terms.

\subsubsection{Partially captured aleatoric error in observed data}\label{sec: partial_aleatoric}

In Supplementary Information Equations \ref{eq:observed_outcome}-\ref{eq:observed_predictors}, we assume that $y_{\text{observed}}$ captures all aleatoric error in $y_{\text{true}}$. However, since $y_{\text{observed}}$ is measured after constructing $y_{\text{true}}$, this assumption may not hold in all situations. For example, observed data may be a biased sub-sample of the whole population. With such observed data, we can only optimize epistemic errors ($\delta_f,\delta_\textbf{x} \text{ and } \delta_y$) that are inherent in training data. In this case -- even when all epistemic error is reduced -- $\varepsilon_{y_{\text{observed}}}$ cannot represent the true $\varepsilon_y$. Consider a true target variable which is measurable across a population ($y_{\text{true}}$), but $y_{\text{observed}}$ only covers a non-representative 70\% of the population. In this case, $\varepsilon_{y_{\text{observed}}}$ only captures the $\varepsilon_y$ of the observed sample, and lacks the ability to capture the component which would be inherent in the remaining 30\% of the population. The representativeness of the error term $\varepsilon_{y_{\text{observed}}}$ depends on how well $y_{\text{observed}}$ represents true reality ($y_{\text{true}}$). It also affects assumptions we need to make in terms of the bias-variance decomposition as per Equations \ref{eq:bias_var}-\ref{eq:bias_var_decomposition}. If $\varepsilon_{y_{\text{observed}}} \neq \varepsilon_y$, the covariances between $\varepsilon_{y_{\text{observed}}}$ and $\delta_f, \delta^{\text{error}}_y \text{ and } \delta^{\text{error}}_\textbf{x}$ are no longer zero. The $\operatorname{Var}(Error)$ term must also incorporate $\operatorname{Cov}(\varepsilon_{y_{\text{observed}}},\delta_f), \operatorname{Cov}(\varepsilon_{y_{\text{observed}}},\delta^{\text{error}}_y), \operatorname{Cov}(\varepsilon_{y_{\text{observed}}},\delta^{\text{error}}_\textbf{x})$. Therefore, guaranteeing the representativeness of $y_{\text{observed}}$ to $y_{\text{true}}$ (which also affects the representativeness of $\textbf{x}_{\text{observed}}$ to $\textbf{x}_{\text{true}}$ from a measurement perspective) is a necessary step before considering epistemic errors in the form of $\delta_f, \delta_y, \text{ and } \delta_\textbf{x}$, as it interrogates whether the questions we are answering are really what we meant to ask. 

\newpage

\section{Supplementary Figures}
\begin{figure}[!h]
\begin{center}
\includegraphics[width=1\textwidth]{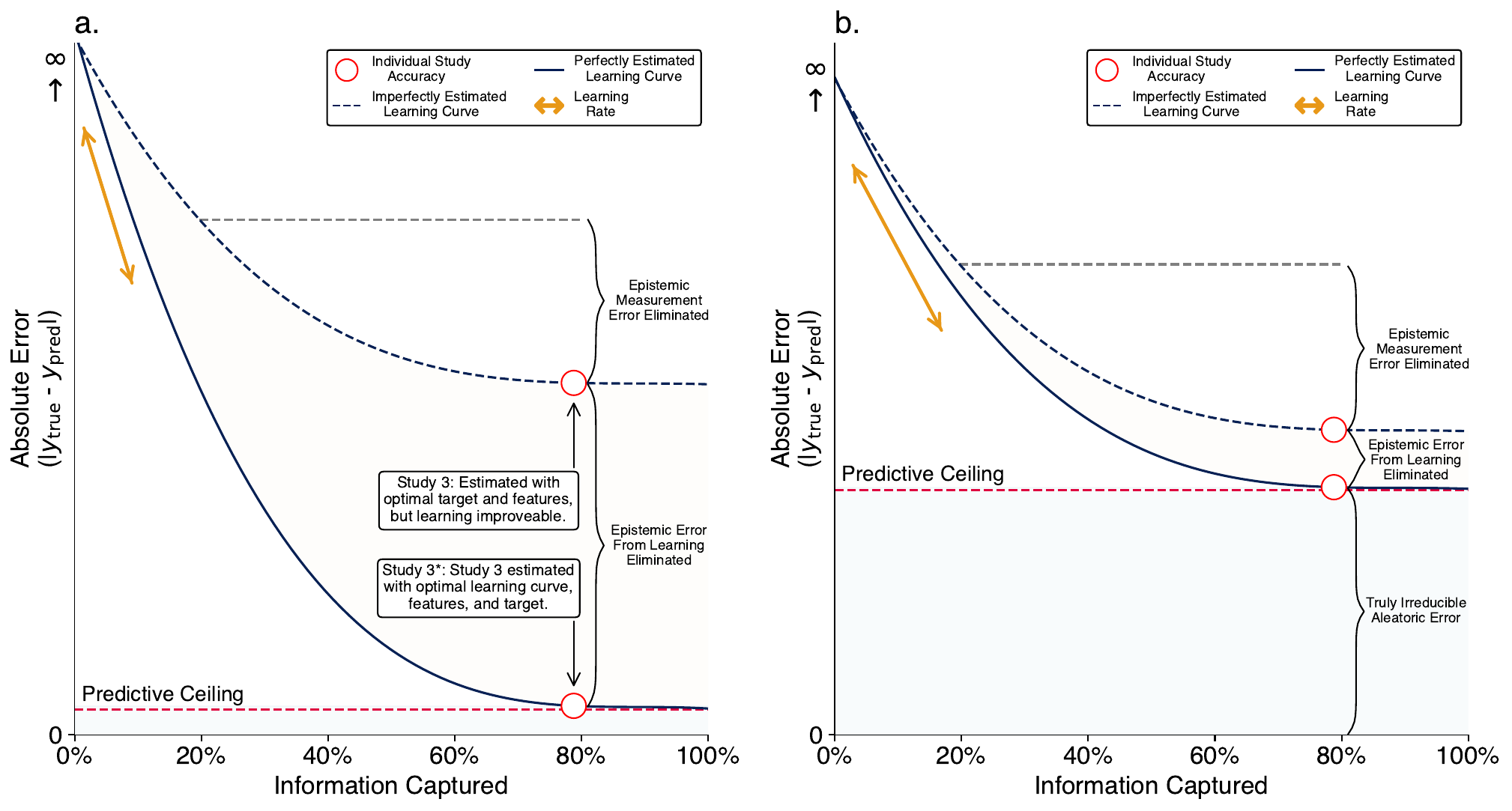}
\captionsetup{labelformat=empty} 
\caption{\textbf{Supplementary Figure 1: Additional Learning Curve Examples.} Two complementary examples to Figure 1. Panel `a.' represents a target variable with relatively little aleatoric error, and a rapid learning rate with large gains from eliminating measurement error. Panel `b.' represents a target variable with relatively higher aleatoric error, a slower learning rate, and relatively fewer gains from improving measurement. Note: there is no requirement for such learning curves to be continuous functions. Dashed horizontal red lines denote what is commonly known as the `predictive ceiling' (no prediction error). Golden arrows denote the rate at which reducible error is reducing. Dashed blue lines represents a less accurate learning algorithm, and the solid blue line represents the case where learning error is entirely eliminated.}
\label{figure_si1}
\end{center}
\end{figure}

\newpage
{
  \renewcommand{\refname}{Supplementary References}

}

\end{bibunit}

\end{document}